\RequirePackage{fix-cm}

\documentclass[twocolumn]{svjour3}          
\smartqed  
\usepackage{graphicx}
\usepackage{natbib}

\usepackage{tabulary,multirow,overpic,xcolor}
\usepackage[misc,geometry]{ifsym} 
\usepackage{authblk}
\usepackage[colorlinks,citecolor=blue,urlcolor=blue,bookmarks=false,hypertexnames=true]{hyperref}
\usepackage[toc,page]{appendix}
\usepackage{arydshln}
\usepackage{amsmath,amsfonts,bm}
\usepackage{url}
\usepackage{mathtools}
\usepackage{amssymb}             
\usepackage{subcaption}
\usepackage{colortbl}
\usepackage{booktabs}
\usepackage{pifont}
\usepackage{diagbox}
\usepackage[T1]{fontenc} 
\usepackage[ruled,lined]{algorithm2e}
\newcommand{\etal}{\textit{et al}.}
\newcommand{\ie}{\textit{i}.\textit{e}.}
\newcommand{\eg}{\textit{e}.\textit{g}.}
\journalname{Preprint}

\begin{document}
\sloppy

\title{Combating Label Noise With A General Surrogate Model For Sample Selection
}

\author{Chao Liang
\and Linchao Zhu
\and Humphrey Shi
\and Yi Yang}

\institute{
$^{\textrm{\Letter}}$ Linchao Zhu$^{1}$ \\
\email{zhulinchao@zju.edu.cn}  \\
\\
Chao Liang$^{1}$ \\
\email{cs.chaoliang@zju.edu.cn}  \\
\\
Humphrey Shi$^{2,3}$ \\
\email{shihonghui3@gmail.com} \\
\\
Yi Yang$^{1}$ \\
\email{yangyics@zju.edu.cn} \\
\at
1 ReLER Lab, CCAI, Zhejiang University \\
2 SHI Labs @ UIUC \& Oregon \\
3 Picsart AI Research (PAIR) \\
}

\date{Received: date / Accepted: date}

\maketitle

\begin{abstract}
Modern deep learning systems are data-hungry. Learning with web data is one of the feasible solutions, but will introduce label noise inevitably, which can hinder the performance of deep neural networks. Sample selection is an effective way to deal with label noise. The key is to separate clean samples based on some criterion. Previous methods pay more attention to the small loss criterion where small-loss samples are regarded as clean ones. Nevertheless, such a strategy relies on the learning dynamics of each data instance. Some noisy samples are still memorized due to frequently occurring corrupted learning patterns. To tackle this problem, a training-free surrogate model is preferred, freeing from the effect of memorization. In this work, we propose to leverage the vision-language surrogate model CLIP to filter noisy samples automatically. CLIP brings external knowledge to facilitate the selection of clean samples with its ability of text-image alignment. Furthermore, a margin adaptive loss is designed to regularize the selection bias introduced by CLIP, providing robustness to label noise. We validate the effectiveness of our proposed method on both real-world and synthetic noisy datasets. Our method achieves significant improvement without CLIP involved during the inference stage.
\end{abstract}

\section{Introduction}
With the emergence of deep neural networks~(DNNs) and the boost of computation capability, current visual intelligence systems can excel in several tasks, \eg, image classification~\citep{russakovsky2015imagenet,he2016deep,dosovitskiy2021an,liu2021swin}, object detection~\citep{carion2020end, liu2020deep}, video understanding~\citep{2022_centerclip,ma2022weakly}, even surpassing human-level performance. These remarkable breakthroughs are closely related to the collection of high-quality annotated data. However, the labeling process is labor-intensive and expensive. For some specific domains like insect classification, it is much more difficult to annotate the data without expert knowledge. 

Some researchers resort to a compromising scheme and make use of large-scale cheap webly annotated data~\citep{xiao2015learning, kolesnikov2020big}.
This can inevitably introduce label noise. Supervised learning often assumes that the training and test data are sampled from the independent identical distribution. The existence of noisy labels results in a discrepancy between the training and test distribution. As a consequence, learning with noisy labels leads to poor generalization on clean unseen test data.

\begin{figure}[t]
	\center
	\includegraphics[width=1.0\linewidth]{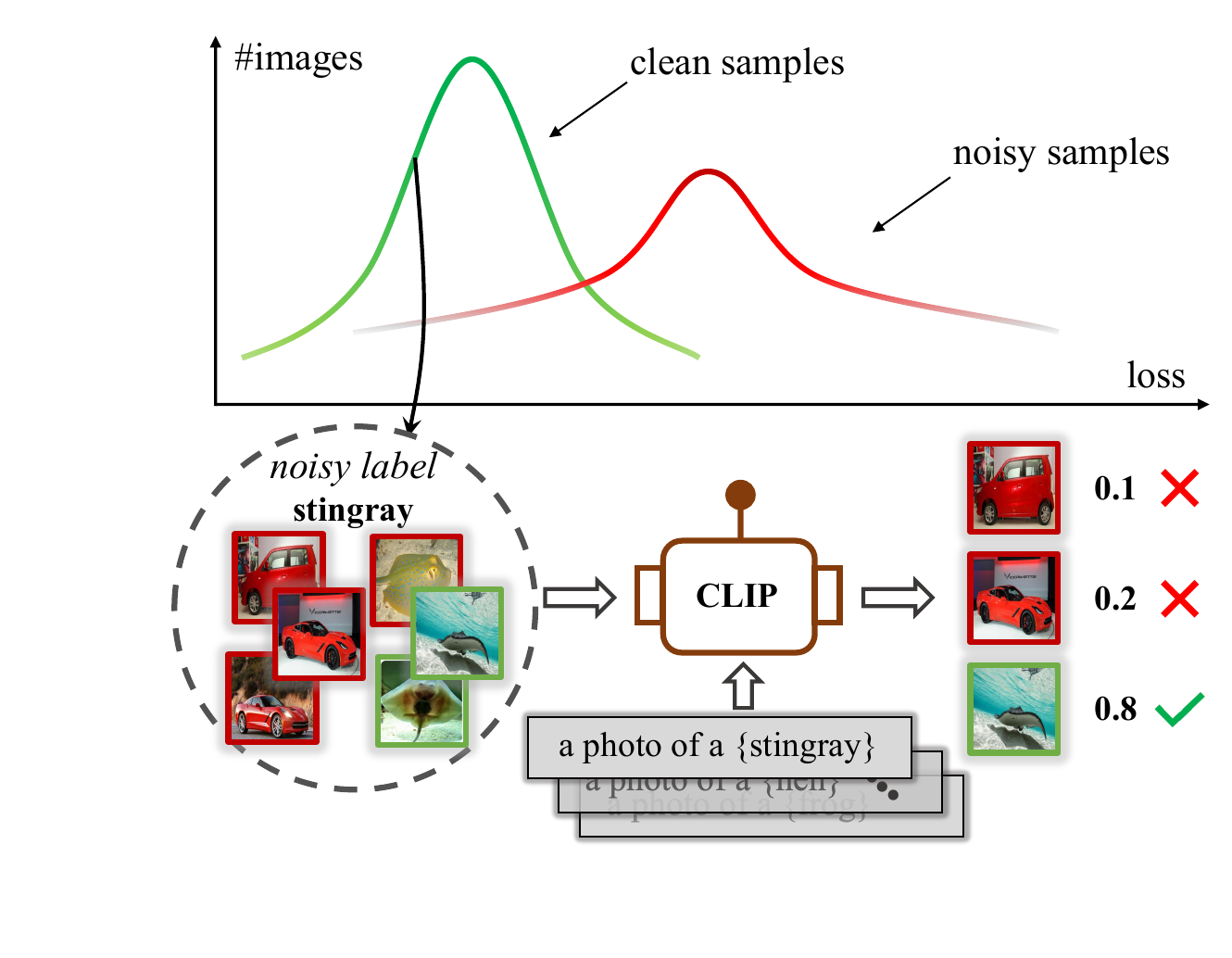}
	\caption{Some small-loss noisy samples that share similar visual patterns are memorized by DNNs. They are misidentified as clean samples by the small loss criterion. With the help of the powerful open-vocabulary vision-language model CLIP, these samples can be further filtered out potentially. With cleaner training samples, the classification performance is further boosted.}
	\label{fig:intro} 
\end{figure}

Massively sophisticated algorithms~\citep{NEURIPS2018_a19744e2, Li2020DivideMix:, wei2020combating, liu2020early, ortego2021multi, xia2021sample, yao2021jo,zhu2022detecting, li2022selective} have been proposed to alleviate the negative effect of noisy labels. One of the promising lines of works is sample selection \citep{NEURIPS2018_a19744e2,wei2020combating, Li2020DivideMix:, yao2021jo, zhu2022detecting}. The main idea is to separate clean samples from all the training samples based on some rules or criteria. The cleaner samples can enhance the learning of a more unbiased classifier. Previous methods~\citep{NEURIPS2018_a19744e2,wei2020combating,Li2020DivideMix:,li2022selective} mostly consider the small loss strategy. Clean and simple samples are assumed to be first fitted by DNNs and noisy samples are later memorized during the training process~\citep{arpit2017closer}.
This strategy heavily relies on the learning dynamics of each data instance and suffers from the undesirable learning bias in the training dataset. 
As shown in Figure~\ref{fig:intro}, a proportion of noisy samples are identified as clean samples with small losses because they share similar corrupted visual patterns that occur frequently in the learning process.
After sample selection, these out-of-distribution noisy samples are still mixed within the pick-out in-distribution clean samples. Learning with these noisy samples can mislead the classifier and have a negative impact on the decision boundary. In order to get rid of the memorization effect, a training-free surrogate model is a good choice for detecting noisy samples.

Recently, vision-language models pretrained on text-image pairs show promising zero-shot performance on downstream tasks, especially CLIP \citep{radford2021learning}. Born with the powerful zero-shot capability, CLIP can be easily adapted to score for unseen objects without extra training. With the text query, it is flexible to take advantage of CLIP to infer the data instance with the correct label or not. 
Benefiting from pretrained on large-scale text-image web data, CLIP shows great robustness to distribution shift and domain generalization. CLIP can bring external knowledge to facilitate the selection of clean samples, which could potentially filter out those noisy samples that have been memorized by DNNs.

In this paper, we leverage the off-the-shelf vision-language surrogate model CLIP~\citep{radford2021learning} to detect noisy samples automatically, which has not been explored yet. \textbf{First}, in contrast to the learning-centric small loss criterion, our CLIP-based selection strategy is a training-free method. Such property avoids the learning bias brought by the noisy supervision. CLIP scores each data instance with its ability of text-image alignment. Combined with the prompt technique, each training sample can be evaluated by CLIP and assigned a surrogate confidence corresponding to its noisy label. Naturally, we regard those samples with high confidences as clean ones. Those noisy samples with corrupted visual patterns can be filtered out with the help of external knowledge, which can further improve the learning of the classifier.
\textbf{Second}, we propose a robust noise-aware balanced margin adaptive loss to regularize the selection bias brought by CLIP.
On the one hand, CLIP remains biased towards certain classes so it can be overconfident in some classes. Also, the existing methods often neglect the side effect of sample selection, that is, the class imbalance issue might occur. Our noise-aware balanced margin adaptive loss modifies the logits directly, encouraging a relatively large margin for overconfident and dominant classes.
This unified margin mechanism can mitigate the effect of noisy labels and imbalanced distribution for robust training.

We evaluate our method on both real-world and synthetic noisy datasets without CLIP involved during the inference stage. The significant improvement on several noisy benchmarks confirms the effectiveness of our proposed method.

Overall, our contributions can be summarized as follows:
\begin{itemize}
	\item We are the first to leverage the off-the-shelf vision-language surrogate model CLIP to help select clean samples automatically, which has not been explored before. This training-free method prevents the learning bias brought by the small-loss strategy, which can improve the learning of a more robust classifier and alleviate the memorization effect.
	\item We propose a noise-aware balanced margin adaptive loss to reduce the selection bias introduced by CLIP, providing much more robustness to label noise.
	\item We demonstrate that our proposed method can achieve significant improvement on both real-world and synthetic noisy datasets without CLIP involved during the inference stage.
\end{itemize}

\section{Related Works}
\label{sec:related}
Numerous approaches have been proposed to combat label noise in recent works~\citep{Li2020DivideMix:, bai2021memomentum, zhang2021learning, zhu2022detecting, huang2023genkl}. The common solutions can be typically categorized into three types: sample selection, sample reweighting, and label correction. 

\noindent\textbf{Sample selection} focuses on identifying the clean samples from all the noisy training samples. The clean samples are then used to train the deep neural network. The key problem is to design a good criterion. There are several strategies~\citep{NEURIPS2018_a19744e2, arazo2019unsupervised, wei2020combating, yao2021jo, ortego2021multi, zhu2022detecting} to detect noisy labels. Among them, the small-loss trick~\citep{NEURIPS2018_a19744e2, arazo2019unsupervised, wei2020combating, yao2021jo} plays an important role. Deep neural networks tend to learn clean and simple patterns faster~\citep{arpit2017closer}. Co-teaching~\citep{NEURIPS2018_a19744e2} selects a pre-defined proportion of samples with small cross-entropy losses and discards the remaining. Instead, JoCoR~\citep{wei2020combating} selects samples with small joint losses composed of cross-entropy losses and co-regularization losses. However, JoSRC~\citep{yao2021jo} argues that prior methods neglect different noise ratios in different mini-batches. It exploits the Jensen-Shannon~(JS) divergence which serves as the sample cleanness, to separate clean samples in a global manner. Recently, several works~\citep{ortego2021multi, zhu2022detecting} try to filter noisy samples out by leveraging neighborhood information, especially via K-Nearest-Neighbors~(KNN) algorithm. MOIT~\citep{ortego2021multi} selects the confident examples based on the representation similarity between the neighbors. Zhu~\etal~\citep{zhu2022detecting} employ KNN to re-label each sample and detect noisy labels by two simple criteria: local majority voting and global score-based ranking.

\noindent\textbf{Sample reweighting} is a traditional and effective method to resist the memorization effect of noisy labels, which encourages larger weights for clean samples and smaller weights for noisy ones~\citep{shu2019meta, zhang2021learning, xu2021faster}. Meta-Weight-Net~\citep{shu2019meta} learns to reweight each sample following the meta-learning paradigm. However, this method requires a small unbiased, and clean validation set, which might be difficult or expensive to collect in practice. To overcome this limitation, Zhang~\etal~\citep{zhang2021learning} propose to build the proxy clean data from the training history. They maintain the memory to store the past losses and use the changes between the model and meta-model at different steps as the selection criterion.

\begin{figure*}[t]
	\center
	\includegraphics[height=0.34\linewidth]{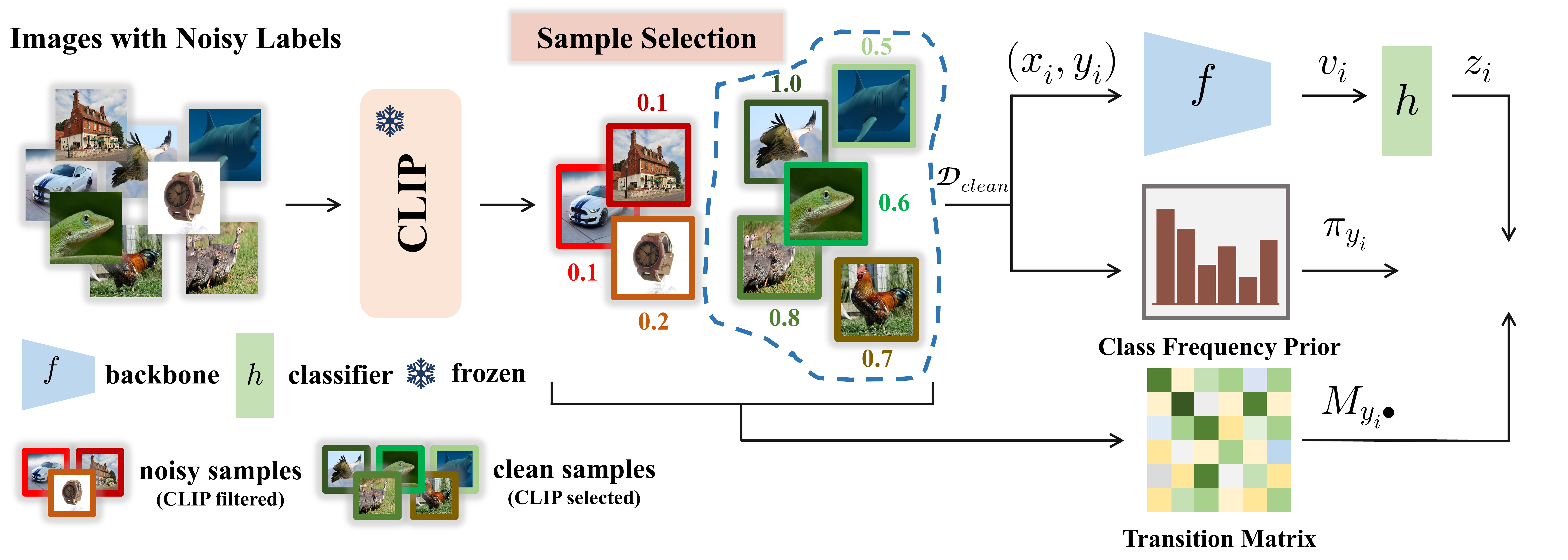}
	\put(-50, 83.9){$\mathcal{L}_{\text{NABM}}$~(Eq.~\ref{eq:nabm_loss})
	}
	\caption{The overall framework is presented. We leverage the open-vocabulary vision-language surrogate model CLIP to select clean samples. The annotated confidence is predicted by CLIP corresponding to its noisy label. Here, \textcolor{red}{red} denotes noisy samples treated by CLIP and \textcolor{green}{green} denotes clean ones. Then, combined with the transition matrix and the class frequency prior, we propose a noise-aware balanced margin adaptive loss to mitigate the overconfidence effect and the class imbalanced issue.}
	\label{fig:framework} 
\end{figure*}

\noindent\textbf{Label correction} aims to assign correct pseudo labels to those samples with wrong labels. The most popular way is to use the prediction from the model~\citep{arazo2019unsupervised, chao2023stitchup}. In general, the generated pseudo label is the convex combination between the original noisy label and the current prediction of the model~\citep{arazo2019unsupervised}. Some works utilize the prediction from class prototypes~\citep{han2019deep} or get hard labels based on threshold~\citep{ortego2021multi}.

Others combine several techniques to prevent overfitting to noisy labels, \eg, mix-up~\citep{zhang2018mixup}, label smoothing~\citep{ortego2021multi}, consistency regularization~\citep{iscen2022learning, cheng2022classdependent}, semi-supervised framework~\citep{Li2020DivideMix:}, contrastive learning~\citep{ortego2021multi, wu2021ngc}. DivideMix~\citep{Li2020DivideMix:} first divides the training samples into the labeled and unlabeled set by fitting the Gaussian Mixture Model~(GMM) on the loss distribution, and then performs the semi-supervised learning. NCR~\citep{iscen2022learning} proposes a consistency regularization term to enforce the output logit of one sample similar to its neighbors based on the structure of the feature space.

LAION~\citep{schuhmann2021laion} and DataComp~\citep{gadre2024datacomp} employ CLIP to filter image-text pairs during the training of vision-language models, with the primary objective of enhancing data quality by eliminating noisy or irrelevant pairs. These approaches leverage CLIP’s multi-modal understanding to ensure that only the most semantically aligned image-text pairs contribute to the training process. While our proposed method shares the fundamental principle of data refinement, it introduces a novel application of CLIP in a different context: filtering noisy labels in classification tasks.

Unlike the prior works, which focus on the relational alignment between image and text pairs in multi-modal datasets, our method specifically addresses the issue of mislabeled data within a purely visual classification setting. Here, the noise is label-centric, where an image is incorrectly labeled, leading to inaccuracies in the training dataset. By applying CLIP to evaluate the consistency between images and their associated labels, we can effectively identify and remove incorrect labels, thereby improving the accuracy and reliability of the labeled dataset. This novel use of CLIP for label noise filtering in classification represents a significant departure from its traditional role in vision-language model training, marking a new frontier in its application for enhancing dataset quality.

\section{Method}

\label{sec:method}

\subsection{Preliminary}
\noindent\textbf{Problem formulation.} In the image classification problem, we are given a training dataset $\mathcal{D} = \{(x_i, y_i) | i=1, 2, 3, ..., N\}$ consisted of $N$ sample pairs, where $x_i$ is an image and $y_i \in \{1, 2, 3, ..., C\}$ is the associated label for each sample pair. $C$ is the number of classes. In our task, some unknown number of labels are noisy, \ie, $y_i \ne \hat{y}_i$, where $y_i$ is the noisy label and $\hat{y}_i$ is its true class label. Note that $y_i$ is the correct label if and only if $y_i = \hat{y}_i$. Our goal is to train a deep neural network $\mathcal{F}_{\theta}$ with such a noisily labeled training dataset, which can generalize well on the clean unseen test data. The network $\mathcal{F}_{\theta}$ is composed of three components: (1) a feature encoder $f$ that maps an image $x_i$ into a high-dimensional representation $v_i = f(x_i)$; (2) a classifier $h$. It takes $v_i$ as an input and outputs the logit $z_i = h(v_i)$; (3) a softmax layer $\sigma$ transforms the logit $z_i$ into the probability $p_i$.

\noindent\textbf{Vision-language surrogate model.} Recently, models pretrained on large-scale text-image supervision have been popular, \eg, CLIP~\citep{radford2021learning}. In the training stage, CLIP pretrains the image encoder and text encoder with the contrastive loss. It pulls the image feature embedding and the paired text feature embedding closer in the shared embedding space by maximizing the cosine similarity. During the inference, CLIP can predict the most possible pair given an image and a set of prompt-based texts like ``a photo of a \{CLASS\}", where \{CLASS\} is replaced by the class name. This framework endows CLIP with the capability of open-vocabulary zero-shot classification naturally, and it can be adapted to several downstream tasks. By the power of text-image alignment, we leverage the CLIP-like open-vocabulary vision-language surrogate model to select clean samples based on the prediction confidences. Note that the frozen CLIP is only used as a scorer in the training stage.

\noindent\textbf{Overview.} First, we pretrain the feature encoder $f$ to learn the robust representation with noisy labels. Then, we only keep the backbone $f$ and re-train the classifier $h$. We apply the vision-language pretrained model CLIP~\citep{radford2021learning} to help select clean samples automatically. In order to mitigate the selection bias introduced by CLIP, we design a robust noise-aware balanced margin adaptive loss to regularize the effect of overconfidence and class imbalance. 
The overall framework is presented in Figure~\ref{fig:framework}.
We present the overall training algorithm in Algorithm~\ref{alg:overall}.

\subsection{Selecting clean samples with CLIP}
\label{subsec:selct_clip}

Learning with noisy labels suffers from the adverse effect that deep neural networks can easily memorize noisy samples~\citep{zhang2021understanding}. One of the effective solutions to this problem is sample selection. Most of the prior research~\citep{NEURIPS2018_a19744e2, wei2020combating, Li2020DivideMix:, li2022selective} relies on the small loss criterion, based on the observation that deep networks fit clean samples first, and then gradually noisy ones~\citep{arpit2017closer}. This strategy is a learning-centric selection metric by fitting the data distribution. It can be affected by the learning bias in the training dataset where those noisy samples with repetitive corrupted visual patterns are identified as clean samples. As a result, the deep network accumulates the prediction errors. To avoid this confirmation bias, we resort to a training-free surrogate model. We leverage the off-the-shelf pretrained surrogate model CLIP~\citep{radford2021learning} to help detect clean samples automatically. CLIP shows several advantages in learning with noisy labels: (1) a training-free selection strategy devoid of reliance on memorization effect; (2) flexible to transfer to downstream tasks with the powerful capability of text-image alignment without extra training; (3) customized prompt engineering that might help filter out some noisy labels based on our prior knowledge potentially. 

We propose to select clean samples based on the predictions from CLIP. Given an images $x$, the image feature $V$ is extracted by the image encoder and the text features $\{T_1, ..., T_C\}$ are generated by the text encoder from the prompt template $\mathcal{T}$, \eg, ``a photo of a \{CLASS\}". Then, the CLIP prediction for label $y=i$ is computed as follows:
\begin{equation}
	\label{eq:clip_pred}
	q(y=i|x) = \frac{\exp(\cos(V, T_i)/\tau)}{\sum_{j=1}^{C}\exp(\cos(V, T_j)/\tau)},
\end{equation}
where $\cos(\cdot, \cdot)$ denotes the cosine similarity and $\tau$ is the temperature factor. We use $\tau = 0.01$ in the experiments. Then, we consider two types of selection criteria.

\noindent\textbf{Prediction Confidence}: Naturally, we regard the prediction corresponding to the noisy label from the CLIP as the confidence of the sample and select those with high confidences. Specifically, given a sample $x_i$ with a label $y_i$, it is judged as a clean sample if $q_i(y=y_i|x_i) > \rho$, where $\rho$ is a pre-defined threshold. This criterion is simple and effective.

\noindent\textbf{Prompt Consistency}: Domain-specific knowledge can be injected into the prompt, which helps detect out-of-domain noisy samples. Noisy web images are collected by keyword searching. However, class names can be ambiguous. For example, ``stingray'' can represent a type of a car or an animal. If we target classifying the animals, these car images are treated as out-of-domain data. It is difficult for the small loss criterion to distinguish these noisy samples because these images share repetitive visual patterns. Models can easily memorize these samples. Prompts help specify the content of the images. The prediction for a clean sample should be consistent between two prompts where the only difference is domain-specific context. For instance, we apply two prompt templates $\mathcal{T}_1$: ``a photo of a \{CLASS\}'' and $\mathcal{T}_2$: ``a photo an animal \{CLASS\}'' to get two predictions $q_i$ and $\tilde{q}_i$ for a given sample $x_i$. We utilize the Jensen-Shannon divergence to quantify the distance $d_i$ between the above two probability predictions:
\begin{align}
	d_i &= D_{JS}(q_i || \tilde{q}_i) \nonumber\\ 
	&= \frac{1}{2} D_{KL}(q_i || \frac{q_i + \tilde{q}_i}{2}) + \frac{1}{2} D_{KL}(\tilde{q}_i || \frac{q_i + \tilde{q}_i}{2}),
\end{align}
where $D_{KL}$ is the Kullback-Leibler~(KL) divergence. 
Intuitively, we treat samples with small JS divergence as clean samples, \ie, $d_i < \mu$, where $\mu$ is a pre-defined threshold. This criterion allows us to make use of human knowledge to help detect noisy samples but it may need sophisticated design.

By introducing external knowledge from CLIP~\citep{radford2021learning, yang2021multiple}, the noisy samples that have been memorized by DNNs can be further identified potentially. The selected cleaner samples can facilitate the learning of a more robust classifier and therefore improve the classification performance. 

\subsection{Noise-Aware Balanced Margin Adaptive Loss}
\label{sec:loss}
CLIP~\citep{radford2021learning} helps select clean samples, nevertheless, it can also bring the selection bias. On the one hand, CLIP is often biased towards some classes~\citep{wang2022debiased}. It can provide overconfident scores for some classes. On the other hand, the class imbalance issue occurs after sample selection, which is often neglected by the existing methods. In order to regularize the selection bias, we take advantage of margin adaptive mechanism with two priors, which encourages the overconfident and dominant classes to have relatively large margins.

\noindent\textbf{Transition matrix}. The transition matrix can be used to reflect the class-level confidence of the model. Each element $M_{ij}$ in the transition matrix $M \in \mathbb{R}^{C\times C}$ represents the probability of being flipped to a label $j$ when given an instance with a label $i$. 
Following GLC~\citep{hendrycks2018using}, we estimate the class-dependent transition matrix by the average of the prediction $q(y=i|x)$~(Eq.~\ref{eq:clip_pred}) from the vision-language surrogate model CLIP:
\begin{equation}
\label{eq:transition_matrix}
	M_{ij} = \frac{1}{N_i}\sum q(y=j|x, y=i),
\end{equation}
where $N_i$ denotes the number of instances in class $i$. Note that we estimate the transition matrix by using all the training samples. Addressing noisy labels with the transition matrix has been extensively studied in the literature~\citep{hendrycks2018using, li2022estimating, cheng2022cvpr}. They mostly use the transition matrix to refine the output probability directly. By contrast, we regard it as a margin penalty to prevent the overconfidence effect.

\noindent\textbf{Class frequency prior}. The class frequency prior measures the distribution of the training data, which is a common statistic used to address the long-tail problem~\citep{menon2021longtail}. It is defined as $\pi_j = N_j'/N'$ where $N', N_j'$ are the number of the training samples and the number of instances in class $j$.

With the transition matrix and class frequency prior, we propose a noise-aware margin adaptive loss to address the above mentioned problems in a unified framework. After the selection of clean samples, we get the clean subset $\mathcal{D}_{clean}$ consisting of $N'$ training samples. For $(x_i, y_i) \in \mathcal{D}_{clean}$, we obtain the model's output of the softmax probability $\hat{p}_i$ as:
\begin{equation}
\label{eq:nabm_term}
	 \hat{p}_i = \frac{\exp((z_i^{y_i} + \delta M_{y_iy_i} + t\log \pi_{y_i})/s)}{\sum_{j=1}^{C} \exp((z_i^j + \delta M_{y_ij} + t\log\pi_{j})/s)},
\end{equation}
where $\delta, t$ control the noise-aware margin and balanced margin, respectively. Here, $s$ is the temperature factor. 

Conventionally, the deep neural network is optimized by empirical risk minimization of the vanilla cross-entropy loss:
\begin{equation}
	\mathcal{L}_\text{ERM} = \mathbb{E}_{\mathcal{D}_{clean}}[\ell_\text{CE}(x, y)] = \frac{1}{N'}\sum_{i=1}^{N'} \ell_\text{CE}(x_i, y_i),
\end{equation}
\begin{equation}
	\ell_\text{CE}(x_i, y_i) = - \log \frac{\exp({z_i^{y_i}})}{\sum_{j=1}^{C} \exp(z_i^j)}.
\end{equation}
However, we find this loss does not perform well in the experiments. We hypothesize that there are two groups of data for each class after sample selection: one is many easy samples distributed at the center of the class and the other is few hard samples distributed near the class boundary. Cross-entropy loss assigns the same weight to each sample.
The imbalance between many easy samples and few hard samples makes it difficult for classifier optimization.
In order to tackle this issue, we employ the focal loss~\citep{lin2017focal}.

Finally, combined with the focal loss~$\ell_\text{FL}$~($\gamma=1.0$), our noise-aware balanced margin adaptive loss is defined as:
\begin{equation}
\label{eq:nabm_loss}
	\mathcal{L}_\text{NABM} = \mathbb{E}_{\mathcal{D}_{clean}}[\ell_\text{FL}(\hat{p})] = \frac{1}{N'}\sum_{i=1}^{N'} \ell_\text{FL}(\hat{p}_i).
\end{equation}

By modifying the logit with the transition matrix and class frequency prior, this margin adaptive mechanism can suppress overconfidence on biased classes and mitigate the negative effect of imbalanced distribution brought by sample selection from CLIP, which can encourage the model to resist label noise better.

{
\begin{algorithm}
\small
\SetKwData{Left}{left}\SetKwData{This}{this}\SetKwData{Up}{up}
\SetKwInOut{Input}{Input}\SetKwInOut{Output}{Output}
\LinesNumbered
\Input{training dataset with noisy labels $\mathcal{D}=\{(x_i, y_i)\}_{i=1}^N$; pretrained feature encoder $f$; classifier $h$; open-vocabulary vision-language model CLIP; texts $T=\{T_j\}_{j=1}^{C}$ generated by the prompt template;  threshold $\rho$; maximum epoch $E$}
\Output{deep neural network $\mathcal{F}_{\theta} = \{f, h\}$}

{\footnotesize\tcp{Selecting clean samples with CLIP}}

$\mathcal{D}_{clean} \leftarrow \varnothing$\;

\ForEach{ $(x_i, y_i) \in \mathcal{D}$} {
Compute $q(y=y_i|x_i)$ by Eq.~\ref{eq:clip_pred} with CLIP and $T$ \;
\If {$q(y=y_i|x_i) > \rho$}
{$\mathcal{D}_{clean} \leftarrow \mathcal{D}_{clean} \cup \{(x_i, y_i)\}$\;}
}

{\footnotesize\tcp{Calculating the transition matrix}}
Compute the transition matrix $M$ by Eq.~\ref{eq:transition_matrix} on $\mathcal{D}$\;

{\footnotesize\tcp{Calculating the class frequency prior}}
Compute the class frequency prior $\pi$ on $\mathcal{D}_{clean}$\;

{\footnotesize\tcp{Fine-tune the network}}
Re-initial the classifier $h$\;
\For {$e = 1,...,E$}
{\While {$k < \text{MaxIter}$}{
 Draw a mini-batch $\mathcal{X}_{e}^k =\{(x_b, y_b)\}_{b=1}^B$ from $\mathcal{D}_{clean}$ \;
 Compute the loss $\mathcal{L}_{\text{NABM}}$ by Eq.~\ref{eq:nabm_loss} with $M$ and $\pi$ on $\mathcal{X}_e^k$ \;
 Calculate the gradients by the loss $\mathcal{L}_{\text{NABM}}$ backpropagation \;
 Optimized by SGD\;
}}

\Return $\mathcal{F}_{\theta} = \{f, h\}$
\caption{{Pseudo-code for our method.}}
\label{alg:overall}
\end{algorithm}
}

\section{Experiments}

\label{sec:experiments}
\subsection{Experiment setup}
\noindent\textbf{Datasets.}
We evaluate the effectiveness of our proposed method on both real-world and synthetic noisy datasets. For \textit{real-world} datasets, we conduct experiments on five benchmarks with noisy labels: Clothing1M, WebVision, Red Mini-ImageNet, CIFAR-10N, and CIFAR-100N. Clothing1M~\citep{xiao2015learning} consists of 1 million training images collected from some online shopping websites where labels are produced by the surrounding texts. The test set contains 10,526 images of 14 classes. Webvision~\citep{li2017webvision} is crawled from the web using 1,000 concepts from ImageNet ILSVRC12~\citep{deng2009imagenet}. Following~\citep{Li2020DivideMix:}, we experiment with the first 50 classes of the Google image subset on WebVision 1.0. The training and validation set contains 65,944 and 2,500 images, respectively. Red Mini-ImageNet~\citep{jiang2020beyond} is a benchmark of
controlled real-world label noise from the web. The dataset contains 100 classes. We experiment with the noise rate of 20\%, 40\%, 60\%, and 80\%. The image size is resized to 32$\times$32 for a fair comparison~\citep{garg2023instance, kim2024learning}. CIFAR-10N and CIFAR-100N~\citep{wei2022learning} are two recently proposed benchmarks with real-world human-annotated noisy labels. The noisy labels are collected from Amazon Mechanical Turk. For CIFAR-10N, each image is annotated with 3 human-annotated labels. We study three types of noisy label sets: (1) Aggregate: the noisy label is aggregated by majority voting; (2) Random $i$~($i=1,2,3$): use the $i$-th annotated label as the noisy label; (3) Worst: use any wrongly annotated label if it exists. For CIFAR-100N, each image is annotated with one noisy fine label and a coarse label. Please refer to~\citep{wei2022learning} and the website\footnote{\scriptsize \url{http://noisylabels.com/}} for more details. For \textit{synthetic} datasets, we manually make label corruption on CIFAR-10 and CIFAR-100~\citep{krizhevsky2009learning}. Both CIFAR-10 and CIFAR-100 contain 50,000 training images and 10,000 test images, with 10 and 100 classes, respectively. The size of images is 32$\times$32. We investigate three types of label noise: symmetric, asymmetric, and instance-dependent. Symmetric noise is generated by randomly replacing clean labels with other possible labels. In this case, some clean labels can be maintained. We constrain the label flipping to be closed under the given label set. Asymmetric noise is injected by replacing labels only in similar classes, \eg, deer$\rightarrow$horse, dog$\leftrightarrow$cat, which is more common in practice. Instance-dependent noise depends on image information. We simulate the experiment environment following~\citep{xia2020part}.

\noindent\textbf{Evaluation metrics.} Top-1 test accuracy is reported in the experiments. For Clothing1M dataset, we select the model that performs best on the validation set. For WebVision, we also report top-5 accuracy. After training on WebVision, we evaluate performance on ImageNet without any finetuning.

\noindent\textbf{Implementation details.} Following~\citep{Li2020DivideMix:, li2022selective}, we perform the same experiment protocol to pretrain the backbone. For WebVision, we use Inception-ResNet V2~\citep{szegedy2017inception} for DivideMix~\citep{Li2020DivideMix:} and ResNet-18 for Sel-CL~\citep{li2022selective}. We reinitialize the classifier and train for 10 epochs with a learning rate of 0.01 and 0.001, respectively. The optimizer is SGD with a weight decay of 0.0001 and the batch size is 64. For Clothing1M, we use ResNet-50 pretrained on ImageNet. Following the previous works~\citep{Li2020DivideMix:}, we sample 1000 mini-batches in each epoch. The training epoch is 80 and the initial learning rate is 0.02. We set $\rho = 0.6$. For Red Mini-ImageNet, CIFAR-10N, CIFAR-100N, CIFAR-10, and CIFAR-100 datasets, we train Pre-Act ResNet-18 for 10 epochs. The weight decay is set as 0.0005 and the batch size is 128. We set $\rho = 0.5$ for Red Mini-ImageNet, CIFAR-10N, and CIFAR-10, $\rho = 0.1$ for CIFAR-100 and CIFAR-100N. We use $s = 1.0, \delta = 0.5, t = 1.0$, ViT-B/16 as the backbone of CLIP for filtering when training ResNet-18, and $s = 0.1, \delta = 0.1, t = 0.01$, ResNet-50 for others. Empirically, we find smaller $s$ is suitable for deeper networks.

\subsection{Results}
\begin{table}[t]
	\centering
	\small
	\resizebox{0.99\linewidth}{!}{
		\begin{tabular}	{l |c|c|c|c }
			\toprule	 	
			\multirow{2}{*}{Method}  & \multicolumn{2}{c|}{WebVision} & \multicolumn{2}{c}{ILSVRC12}\\
			\cmidrule{2-5}
			& top-1 & top-5& top-1 & top-5\\
			\midrule			
			F-correction~\citep{patrini2017making} & 61.12 & 82.68 & 57.36 & 82.36\\		
			Decoupling~\citep{malach2017decoupling} & 62.54 & 84.74 & 58.26 & 82.26\\
			MentorNet~\citep{jiang2018mentornet}  & 63.00 & 81.40 & 57.80 & 79.92\\	
			
			Co-teaching~\citep{NEURIPS2018_a19744e2} & 63.58 & 85.20 & 61.48 & 84.70\\			
			Iterative-CV~\citep{chen2019understanding} &  65.24 & 85.34 &  61.60 & 84.98\\	
			ELR~\citep{liu2020early} & 76.26 & 91.26 & 68.71 & 87.84\\
			ELR+~\citep{liu2020early} & 77.78 & 91.68 & 70.29 & 89.76\\
			NGC~\citep{wu2021ngc} & 79.16 & 91.84 & 74.44 & 91.04\\
                TCL~\citep{huang2023twin} & 79.10 & 92.30 & 75.40 & 92.40 \\
                LSL~\citep{kim2024learning} & \textbf{81.40} & 93.00 & \textbf{77.00} & 91.84 \\
                \midrule
                CLIP zero-shot~(RN50)~\citep{radford2021learning} & 71.20 & 95.04 & 74.04 & 96.80\\
                CLIP zero-shot~(ViT-B/16)~\citep{radford2021learning} & 75.44 & 96.92 & 78.36 & 97.92\\
			\midrule
			DivideMix~\citep{Li2020DivideMix:} & 77.32 & 91.64 & 75.20 & 90.84\\
			Ours~(DivideMix init.) & 79.08 & 91.96 & 76.04 & 93.12\\
                \midrule
			Sel-CL~\citep{li2022selective} & 77.88 & 91.60 & 74.28 & 90.96\\
			Ours~(Sel-CL init.) & \underline{81.00} & \textbf{93.84} & \underline{76.28} & \textbf{94.92}\\
			\bottomrule
	\end{tabular}}
        \caption
	{
		We report top-1 and top-5 test accuracy~(\%) on WebVision 1.0 and ImageNet ILSVRC12. Our method achieves significant improvement over baseline methods. We show the \textbf{best} and the \underline{second best} results in LNL methods.
	}
\label{tb:WebVision}
\end{table}	

\begin{table}[t]
\centering
\resizebox{1.0\linewidth}{!}{
\begin{tabular} {l |c|c|c|c }
		\toprule        
		\multirow{2}{*}{Method}  & \multicolumn{4}{c}{Noise rate}\\
		\cmidrule{2-5}
		& 20\% & 40\% & {60\%} & {80\%}\\
		\midrule            
		{CE~\citep{yao2021instance}} & {47.36} & {42.70} & {37.30} & {29.76}\\
            {Mixup~\citep{zhang2018mixup}} & {49.10} & {46.40} & {40.58} & {33.58} \\
            {MentorMix~\citep{jiang2020beyond}} & {51.02} & {47.14} & {43.80} & {33.46} \\
            {DivideMix~\citep{Li2020DivideMix:}} & {50.96} & {46.72} & {43.14} & {34.50} \\
            {SSR~\citep{feng2021ssr}} & {52.18} & {48.96} & {42.42} & {33.20} \\
            {FaMUS~\citep{xu2021faster}} & {51.42} & {48.06} & {45.10} & {35.50} \\
            {LSL~\citep{kim2024learning}} & {54.68} & {49.80} & {45.46} & {36.78} \\
            \midrule
            {InstanceGM~\citep{garg2023instance}}& {58.38} & {52.24} & {47.96} & {39.62} \\
            {Ours$^\dagger$} & {\textbf{61.26}} & {\textbf{57.09}} & {\textbf{53.25}} & {\textbf{45.65}}\\
		\bottomrule
\end{tabular}}
\caption
{
	{Test accuracy~(\%) on Red Mini-ImageNet~(CNWL). Based on InstanceGM, ours$^\dagger$ achieves significant improvement.}
}
\label{tb:cnwl}
\end{table}

\begin{table*}[t]
\scriptsize
\centering
\resizebox{1.0\linewidth}{!}{
	\begin{tabular}{cccccccc|c}
		\toprule
		CE & Co-teaching~\citep{NEURIPS2018_a19744e2}& ELR~\citep{liu2020early} & NCR~\citep{iscen2022learning} & DivideMix~\citep{Li2020DivideMix:} & {Ours} \\
		\midrule
		68.94  & 69.21 & 72.87 & 74.6  & 74.76 & {\textbf{74.84$\pm$0.03}}\\
		\bottomrule
\end{tabular}}
\caption{Comparison between our method and the existing baseline methods on the Clothing-1M dataset. Test accuracy~(\%) are reported.}
\label{tb:Clothing-1M}
\end{table*}

\setlength{\tabcolsep}{3pt}
\begin{table*}[t]
	\small
	\centering
	\begin{tabular}{l|c c c c c| c }
		\toprule
		Dataset & \multicolumn{5}{ c |}{CIFAR-10N} & CIFAR-100N \\
		\midrule
		Types & Aggregate & Random1 & Random2 & Random3 & Worst & Noisy \\
		\midrule
		CE$^*$~(Standard) &  87.22 & 81.59 & 82.22 & 82.06  & 67.45 & 47.54\\
		\midrule
		DivideMix$^*$~\citep{Li2020DivideMix:} & 95.33 & 95.35 & 95.01 & 95.18 & 92.47 & 69.84 \\
		{Ours$^\dagger$} & {\textbf{95.95 $\pm$ 0.05}} & {\textbf{96.17 $\pm$ 0.10}} & {\textbf{95.58 $\pm$ 0.10}} & {\textbf{95.95 $\pm$ 0.05}} & {\textbf{93.67 $\pm$ 0.10}} & {\textbf{72.46 $\pm$ 0.41}}\\
		\bottomrule
	\end{tabular}
	\caption{Test accuracy(\%) on CIFAR-10N and CIFAR-100N. All results use the PreAct ResNet-18 architecture. $^*$We reproduce all the baselines. Ours$^\dagger$ achieves significant improvement against DivideMix. } 
	\label{tb:cifarN}
\end{table*}

\setlength{\tabcolsep}{7pt}
\begin{table*}[t]
	\small
	\centering
	\begin{tabular}{l|c c c c | c |c c c c}
		\toprule
		Dataset & \multicolumn{5}{ c |}{CIFAR-10} & \multicolumn{4}{ c }{CIFAR-100} \\
		\midrule
		Noise type & \multicolumn{4}{ c |}{Sym.} & Asym. & \multicolumn{4}{ c }{Sym.} \\
		\midrule
		Noise rate & 20\% & 50\% & 80\% & 90\% & 40\% & 20\% & 50\% & 80\% & 90\% \\
		\midrule
		Standard & 83.9 & 58.5 & 25.9 & 17.3  & 77.3 & 61.5  & 37.4 & 10.4 & 4.1 \\
		F-correction~\citep{patrini2017making}&83.1&59.4 &26.2 & 18.8 & 83.1  & 61.4& 37.3& 9.0&3.4\\
		Co-teaching+~\citep{yu2019does}& 88.2 & 84.1& 45.5& 30.1& - & 64.1 & 45.3& 15.5& 8.8  \\
		Mixup~\citep{zhang2018mixup} & 92.3& 77.6&46.7 &43.9 & - & 66.0&46.6 &17.6 &8.1\\
		P-correction~\citep{yi2019probabilistic}& 92.0 & 88.7&76.5 &  58.2&88.1 &  68.1& 56.4&20.7 &  8.8\\						
		M-correction~\citep{arazo2019unsupervised}& 93.8&91.9 &86.6 & 68.7&86.3	& 73.4&65.4 &47.6 &20.5	\\
		MOIT+~\citep{ortego2021multi} & 94.1  & -     & 75.8  & -     & 93.2  & 75.9 & -  & 51.4  & -  \\
		ELR+~\citep{liu2020early}        & 94.9  & 93.9  & 90.9  & 74.5  & 88.9 & 76.3 & 72.0  & 57.2  & 30.9 \\
		NCR~\citep{iscen2022learning}   & 95.2   & 94.3  & 91.6  & 75.1  & 90.7 & 76.6 & 72.5  & 58.0 & 30.8 \\  
		NGC~\citep{wu2021ngc} & 95.9 & 94.5 & 91.6 & 80.5 & 90.6 & 79.3 & 75.9 & 62.7 & 29.8\\
		\midrule
		DivideMix~\citep{Li2020DivideMix:}   & 95.7  & 94.4 & 92.9 & 75.4  & 92.1  & 76.9  & 74.2 & 59.6 & 31.0 \\
		Ours$^\dagger$ & \textbf{96.6} & \textbf{95.6} & \textbf{94.1} & \textbf{89.2} & \textbf{95.1} & \textbf{80.3} & \textbf{76.6} & \textbf{63.4} & \textbf{45.7}\\
		\bottomrule
	\end{tabular}
	\caption{Test accuracy(\%) on CIFAR-10 and CIFAR-100 under different noise rates. Sym. is the symmetric noise and Asym. denotes the asymmetric noise. Based on DivideMix~\citep{Li2020DivideMix:}, ours$^\dagger$ achieves significant improvement against DivideMix.} 
	\label{tb:cifar}
\end{table*}

\noindent\textbf{Real-world datasets.} Table~\ref{tb:WebVision} shows the results on WebVision~\citep{li2017webvision}. Our method achieves significant improvement against DivideMix baseline~\citep{Li2020DivideMix:}. The top-1 and top-5 accuracy is 79.08\% and 91.96\% on WebVision validation set, respectively. The performance gains are 1.76\% and 0.32\% compared with DivideMix. When transferred to ImageNet~\citep{deng2009imagenet}, the performance on top-5 accuracy is substantially boosted. Ours achieves 93.12\% accuracy, surpassing the DivideMix method by 2.28\%. With much more robust learned representation pretrained with contrastive learning~\citep{li2022selective}, our method achieves the second best top-1 accuracy 81.00\%. We find our method shows better top-5 test accuracy compared to the state-of-the-art method LSL~\citep{kim2024learning}. These results verify the effectiveness of our proposed sampling strategy and margin mechanism. Compared with other approaches, our method still shows competitive performance, which is effective to resist label noise. {In Table~\ref{tb:cnwl}, we show the test accuracy on Red Mini-ImageNet~\citep{jiang2020beyond} with controllable realistic label noise. Our method outperforms the contemporary methods, which surpasses the second best by 2.9\%, 4.8\%, 5.3\%, and 6.0\% under the noise rate of 20\%, 40\%, 60\%, and 80\% respectively.}
In Table~\ref{tb:Clothing-1M}, we present the comparison between previous methods on Clothing-1M~\citep{xiao2015learning} with realistic noisy labels. Benefiting from training with the selected cleaner samples, our proposed approach achieves consistent improvement over DivideMix~\citep{Li2020DivideMix:}, showing competitive performance. Compared to other methods like NCR~\citep{iscen2022learning}, ELR~\citep{liu2020early}, our method achieves better performance as well. 
The evaluation results on CIFAR-10N and CIFAR-100N are shown in Table~\ref{tb:cifarN}. We can observe that DivideMix~\citep{Li2020DivideMix:} outperforms the basic CE baseline by a large margin. This confirms the robustness of DivideMix. In addition, our proposed method can improve the DivideMix further. These results indicate our method is effective to mitigate the negative effect of real-world noise.

\begin{table}[t]
\centering
\resizebox{1.0\linewidth}{!}{
\begin{tabular} {l |c|c|c|c|c }
		\toprule        
		\multirow{2}{*}{Method}  & \multicolumn{5}{c}{Noise rate}\\
		\cmidrule{2-6}
		& 20\% & 30\% & 40\% & 45\% & 50\%\\
		\midrule            
		{CE~\citep{yao2021instance}} & {30.42} & {24.15} & {21.45} & {15.23} & {14.42}\\
            {Mixup~\citep{zhang2018mixup}} & {32.92} & {29.76} & {25.92} & {23.13} & {21.31}  \\
            {F-correction~\citep{patrini2017making}} & {36.38} & {33.17} & {26.75} & {21.93} & {19.27} \\
            {Reweight~\citep{liu2015classification}} & {36.73} & {31.91} & {28.39} & {24.12} & {20.23} \\
            {Decoupling~\citep{malach2017decoupling}} & {36.53} & {30.93} & {27.85} & {23.81} & {19.59} \\
            {Co-teaching~\citep{han2018co}} & {37.96} & {33.43} & {28.04} & {25.60} & {23.97} \\
            {MentorNet~\citep{jiang2018mentornet}} & {38.91} & {34.23} & {31.89} & {27.53} & {24.15} \\
            {DivideMix~\citep{Li2020DivideMix:}} & {77.07} & {76.33} & {70.80} & {57.78} & {58.61} \\
            {SSR~\citep{feng2021ssr}} & {78.84} & {78.60} & {76.95} & {74.98} & {72.83} \\
            {LSL~\citep{kim2024learning}} & {80.94} & {79.90} & {78.60} & {78.08} & {77.95} \\
            \midrule
            {InstanceGM~\citep{garg2023instance}} & {79.69} & {79.21} & {78.47} & {77.49} & {77.19} \\
            {Ours$^\dagger$} & {\textbf{80.97}} & {\textbf{80.42}} & {\textbf{79.68}} & {\textbf{79.39}} & {\textbf{78.74}}\\
		\bottomrule
\end{tabular}}
\caption
{
	Test accuracy~(\%) on CIFAR-100 with instance-dependent noise. Based on InstanceGM, ours$^\dagger$ achieves significant improvement.
}
\label{tb:cifar-idn}
\end{table}

\noindent\textbf{Synthetic datasets.}
We show the comparison between our method and the previous methods on the manually corrupted dataset CIFAR-10 and CIFAR-100~\citep{krizhevsky2009learning} under different noise rates and various noise conditions in Table~\ref{tb:cifar} and Table~\ref{tb:cifar-idn}. Our method shows better evaluation performance against DivideMix across all the noise rates, achieving state-of-the-art results in almost all the settings. Especially under a high noise rate of ~90\%, our method considerably outperforms DivideMix~\citep{Li2020DivideMix:} by a large margin. The test accuracy is 89.2\% and 45.7\% on CIFAR-10 and CIFAR-100, respectively.
The gap is 13.8\% and 14.7\% against DivideMix, indicating that our method can identify cleaner samples even under the extreme noise rate. When it comes to asymmetric noise, our method achieves 3.0\% improvement versus DivideMix and even surpasses MOIT+~\citep{ortego2021multi} by 1.9\%. Our method is robust to instance-dependent noise as well. Based on InstanceGM~\citep{garg2023instance}, ours achieves better test accuracy than the state-of-the-art method LSL~\citep{kim2024learning}. The gap is 1.0\% under the noise rate of 40\%. The promising results demonstrate CLIP can be easily adapted to help select clean samples. Equipped with our noise-aware balanced margin adaptive loss, the classifier learning is more robust and unbiased.

\subsection{Ablation Study}

\begin{figure}
	\centering
	\includegraphics[width=1.0\linewidth]{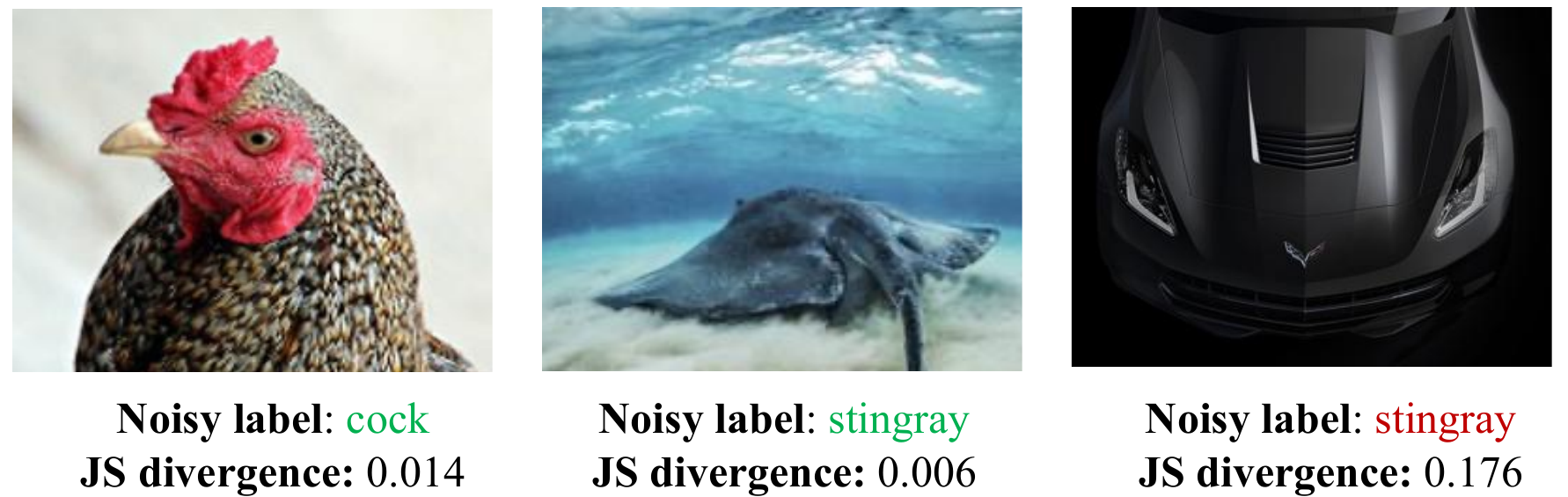}
        \caption{Examples of the selected or filtered images by the prompt consistency strategy. \textcolor[RGB]{192,0,0}{red} denotes the annotated noisy label is wrong and \textcolor[RGB]{0,176,80}{green} represents the annotated noisy label is consistent with its true label.}
	\label{fig:vis_prompt_consistency}
\end{figure}

\noindent\textbf{Sampling strategy.} We explore the effectiveness of our sampling strategy in this ablation analysis. Table~\ref{ab:samplingstrategy} shows the accuracy performance comparison between different sampling strategies. We keep the same experimental setup. The classifier is trained with the focal loss. First, we compare with GMM which is the commonly used method to select clean samples~\citep{arazo2019unsupervised, Li2020DivideMix:}. When the classifier is re-trained with the clean samples that are divided by GMM, the top-1 accuracy on WebVision~\citep{li2017webvision} increases a little compared to the pretrained DivideMix~\citep{Li2020DivideMix:} model. It indicates that the deep neural networks have been memorizing some small-loss noisy samples. These noisy samples cannot be identified by GMM, which can still hinder the learning of a robust classifier. Hence, the performance is hard to further boost. Second, we observe that our prediction confidence strategy significantly improves the classification accuracy, exceeding the baseline by 1\% top-1 accuracy on WebVision. 
\begin{table}[ht]
	\centering
	\small
	\resizebox{1.0\linewidth}{!}{
		\begin{tabular} {l |c|c|c|c }
			\toprule        
			\multirow{2}{*}{Sampling Strategy}  & \multicolumn{2}{c|}{WebVision} & \multicolumn{2}{c}{ILSVRC12}\\
			\cmidrule{2-5}
			& top-1 & top-5& top-1 & top-5\\
			\midrule            
			DivideMix~(pretrained)~\citep{Li2020DivideMix:} & 77.32 & \textbf{91.64} & 75.20 & 90.84\\
			Small Loss~(GMM) & 77.36 & 91.08 & 74.32 & 92.08 \\
			Prediction Confidence & \textbf{78.32} & 91.48 & \textbf{75.28} & \textbf{92.92}\\
			Prompt Consistency &77.36 &91.24 & 74.16 & 92.04 \\
			\bottomrule
	\end{tabular}}
	\caption{
		Comparison between different sampling strategies. We report top-1~(top-5) accuracy~(\%) on WebVision and ImageNet.
	}
\label{ab:samplingstrategy}
\end{table}
The results suggest that CLIP~\citep{radford2021learning} can help detect the memorized noisy samples and select cleaner training samples, with its powerful zero-shot prediction capability and external knowledge from large-scale pretrained data. Learning with cleaner samples contributes to establishing a better decision boundary. Third, we try to inject prior knowledge into the prompt with the expectation that this strategy can identify some out-of-domain data. We find our prompt consistency strategy can help detect noisy samples and select clean ones as shown in Figure~\ref{fig:vis_prompt_consistency}. But the overall performance is almost comparable against GMM. As discussed in Section~\ref{subsec:selct_clip}, we suspect it might require more sophisticated designed prompt templates. We leave it for future work.

\noindent\textbf{Cross-entropy loss v.s. Focal loss.} The straightforward way to address the image classification is to train a classifier with the vanilla cross-entropy loss. However, the performance is unsatisfactory in the experiments. We conduct the experiments on the clean samples selected by CLIP~\citep{radford2021learning}. The results are reported in Table~\ref{ab:loss}. There are some findings: (1) After the selection of clean samples, training with the cross-entropy loss can further boost the top-1 accuracy~(+0.44\%) on WebVision validation set compared to the pretrained model with DivideMix~\citep{Li2020DivideMix:}. This confirms the effectiveness of our sampling strategy. (2) In contrast, the top-1 accuracy on ImageNet drops 0.6\%. As we discussed in Section~\ref{sec:loss}, we hypothesize this is because there are many clean samples with easy patterns in the selected samples. Cross-entropy loss treats each sample equally. DNNs focus more on the simple samples and fit the training data well, which can result in poor performance when transferred to other datasets. (3) Focal loss~\citep{lin2017focal} shows significant improvement against cross-entropy loss on both WebVision and ImageNet, by 0.56\%(1.04\%) and 0.48\%(1.28\%) increase of top-1(top-5) accuracy, respectively. Hard samples play an important role in determining an accurate decision boundary. By balancing the contributions between easy samples and hard samples, focal loss assigns larger weights to hard samples, which prevents the overwhelm of easy samples. Ultimately, it can facilitate the optimization of the classifier and promote the prediction performance.

\begin{table}[ht]
\centering
\resizebox{1.0\linewidth}{!}{
	\begin{tabular} {l |c|c|c|c }
		\toprule        
		\multirow{2}{*}{Loss function}  & \multicolumn{2}{c|}{WebVision} & \multicolumn{2}{c}{ILSVRC12}\\
		\cmidrule{2-5}
		& top1 & top5& top1 & top5\\
		\midrule            
		DivideMix~(pretrained)~\citep{Li2020DivideMix:} & 77.32 & 91.64 & 75.20 & 90.84\\
		Cross-Entropy & 77.76 & 90.44 & 74.80 & 91.64 \\
		Focal Loss & 78.32 & 91.48 & 75.28 & 92.92\\
		\midrule
            {w/o Focal Loss} & {78.20} & {91.04} & {74.92} & {92.00} \\
		w/o balanced margin~($t=0$) & 78.68 & 91.72 & 75.68 & 93.00 \\
		w/o noise-aware margin~($\delta=0$) & 78.76 & 91.92 & 75.76 & 92.92\\
		Ours & \textbf{79.08} & \textbf{91.96} & \textbf{76.04} & \textbf{93.12}\\
		\bottomrule
\end{tabular}}
\caption
{
	Ablation analysis on the loss function and the contribution of each component. We report top-1 and top-5 test accuracy~(\%).
}
\label{ab:loss}
\end{table}

\begin{figure}[t]
	\centering
	\subfloat[\label{fig:ab_threshold}]{
		\includegraphics[width=0.50\linewidth]{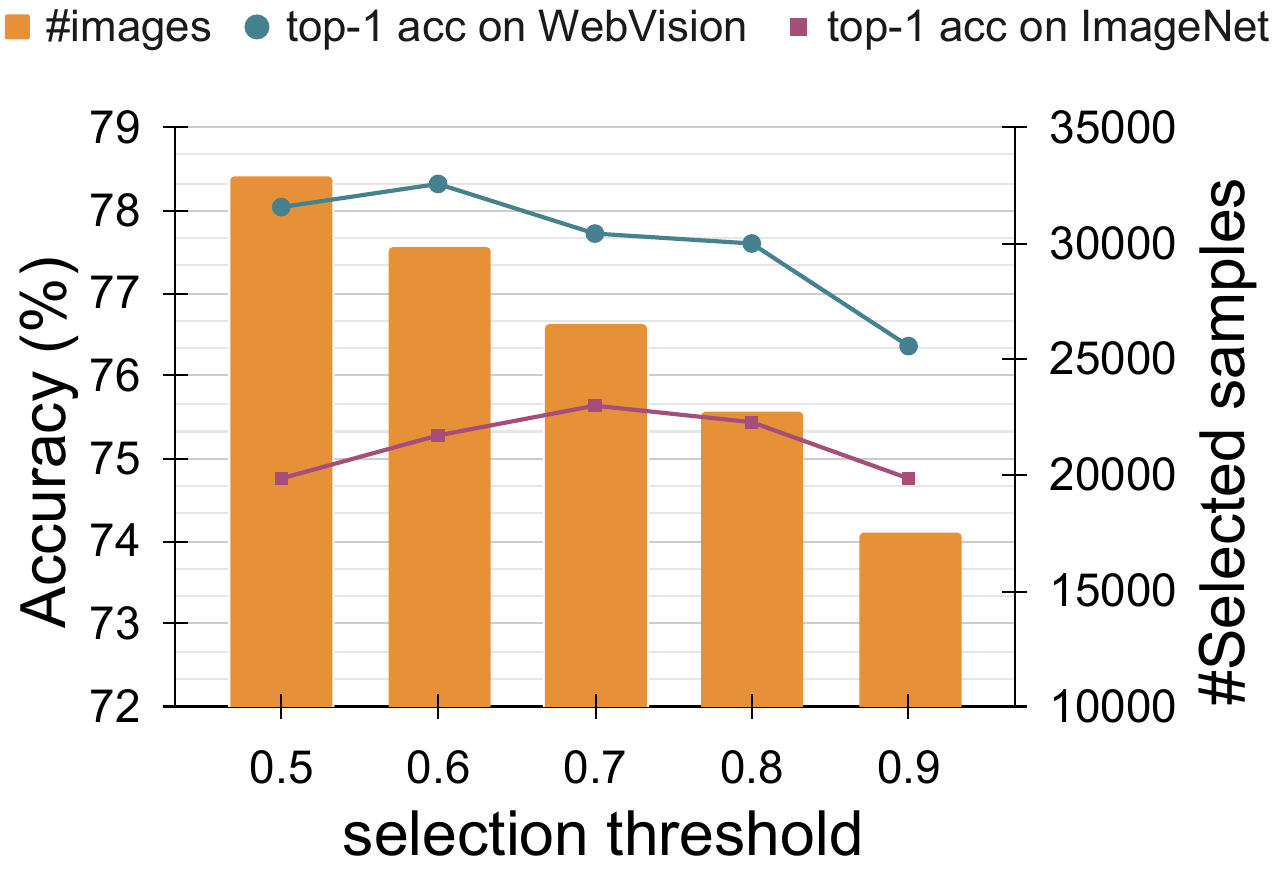}}
	\subfloat[\label{fig:ab_margin}]{
		\includegraphics[width=0.48\linewidth]{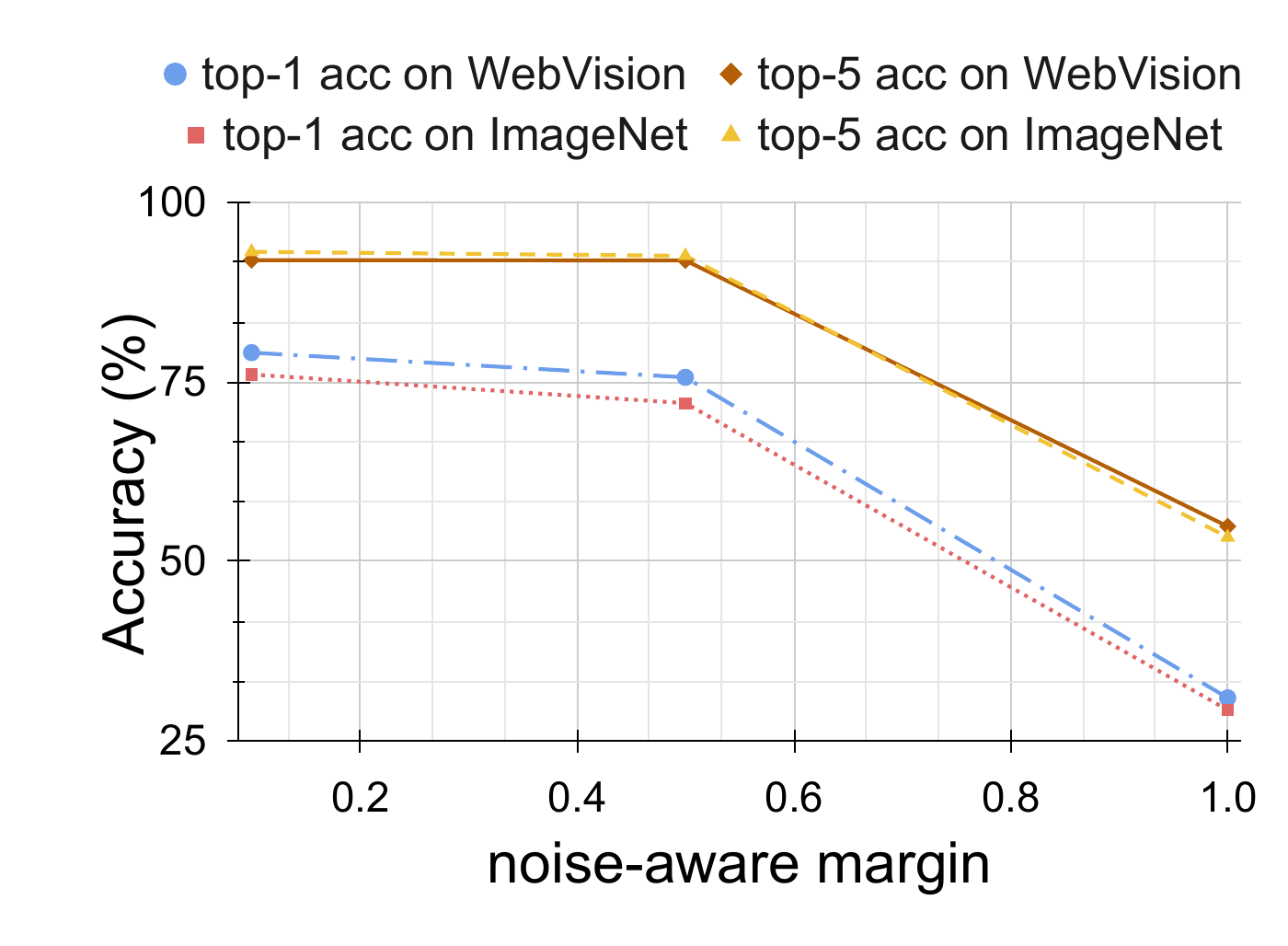}}
	\caption{\ref{fig:ab_threshold}: Ablation study on the effect of selection threshold. We plot the number of training samples and top-1 accuracy~(\%) on WebVision and ImageNet with different thresholds. \ref{fig:ab_margin}: Ablation study on the effect of the noise-aware margin. We report top-1 and top-5 test accuracy~(\%) on both WebVision and ImageNet.}
	\label{fig:ab_threshold_margin}
\end{figure}

\noindent{\textbf{Ablation study on noise-aware balanced margin adaptive loss.} We investigate the contribution of each component in our proposed loss function. For a fair comparison, we conduct the experiments by removing each component and examining the effect. All the other configurations are the same. We present the evaluation results in Table~\ref{ab:loss}. Removing the focal loss hurts the performance. Both noise-aware margin and balanced margin can bring the improvement on the performance. The former outperforms the focal loss baseline by 0.36\% on WebVision~\citep{li2017webvision} and 0.4\% on ImageNet~\citep{deng2009imagenet} measured by top-1 accuracy while the latter obtains the performance gains over 0.44\% and 0.48\%. The observations validate that CLIP~\citep{radford2021learning} model might introduce the selection bias. On the one hand, CLIP can be overconfident in certain classes so that some noisy samples are still mixed even after sample selection. On the other hand, the selected samples often exhibit a class-imbalanced distribution, especially under the situation where asymmetric noise exists. Although sample selection brings cleaner data, the impact of data imbalance will be amplified. Our margin adaptive loss solves these issues in a unified framework, which mitigates the overconfidence effect by the noise-aware margin and relieves the influence of the long-tailed distribution by the balanced margin. The joint effect of these two margins fosters the learning of the classifier and further improves the performance.}

\begin{figure}
	\centering
	\includegraphics[width=1.0\linewidth]{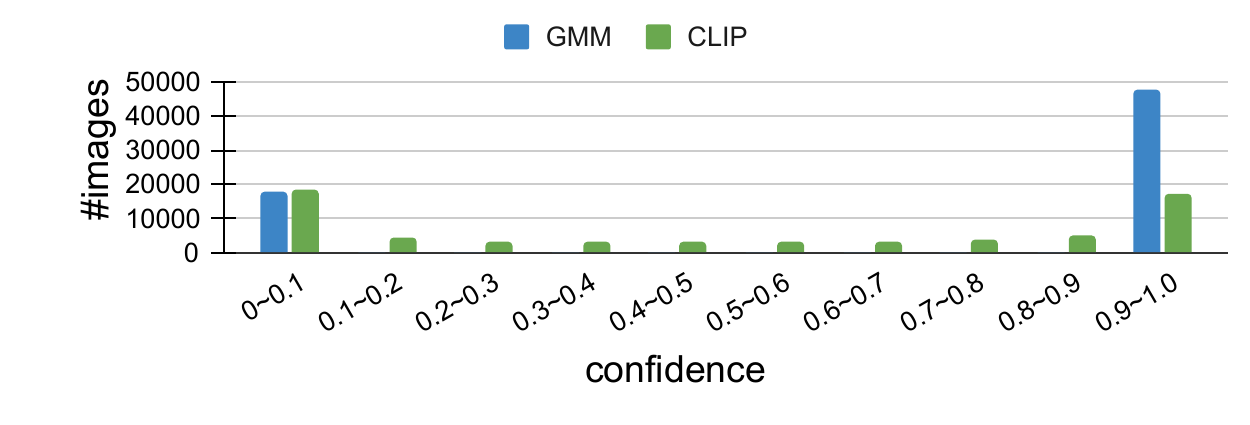}
        \caption{Confidence distribution comparison between GMM and CLIP on WebVision.  We divide confidence into 10 intervals, each with a range of 0.1, and count the image amount for each interval.}
	\label{fig:distribution}
\end{figure}

\begin{figure}
	\centering
	\includegraphics[width=1.0\linewidth]{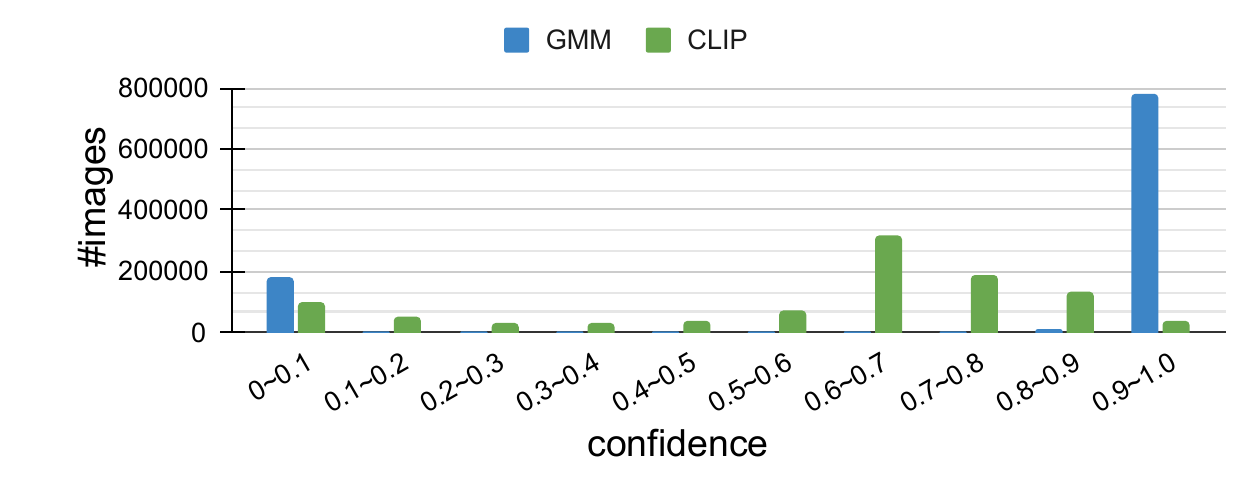}
	\caption{Confidence distribution comparison between GMM and CLIP on Clothing1M. We divide confidence into 10 intervals, each with a range of 0.1, and count the image amount for each interval. }
	\label{fig:clothing_distribution}
\end{figure}

\noindent\textbf{Analysis of transition matrix.} We calculate the error between our estimated and the groundtruth transition matrix. Under the symmetric noise of 0.5, the absolute mean error is 0.02 on CIFAR-10 and 0.006 on CIFAR-100, respectively.

\noindent\textbf{Effect of selection threshold $\rho$ in sample selection.} In Figure~\ref{fig:ab_threshold}, we show the number of the selected training samples and top-1 accuracy on both WebVision~\citep{li2017webvision} and ImageNet~\citep{deng2009imagenet} with the varied selection thresholds on confidence. As we can see, with the threshold getting larger, the number of training samples decreases rapidly. The top-1 accuracy performance on WebVision also drops, especially when we set the threshold to 0.9. This is because the training samples are insufficient to learn a good decision boundary after the selection of the clean samples, especially for some classes with few data points. Even though sample selection can bring cleaner data, we cannot ignore the risk of the reduced amount of training data. Therefore, we choose a proper threshold to ensure enough instances for training the network. We notice that the performance on ImageNet is less affected. We guess it might be attributed to the robust learned representation by training with diverse web data.

\begin{figure}
	\centering
	\includegraphics[width=1.0\linewidth]{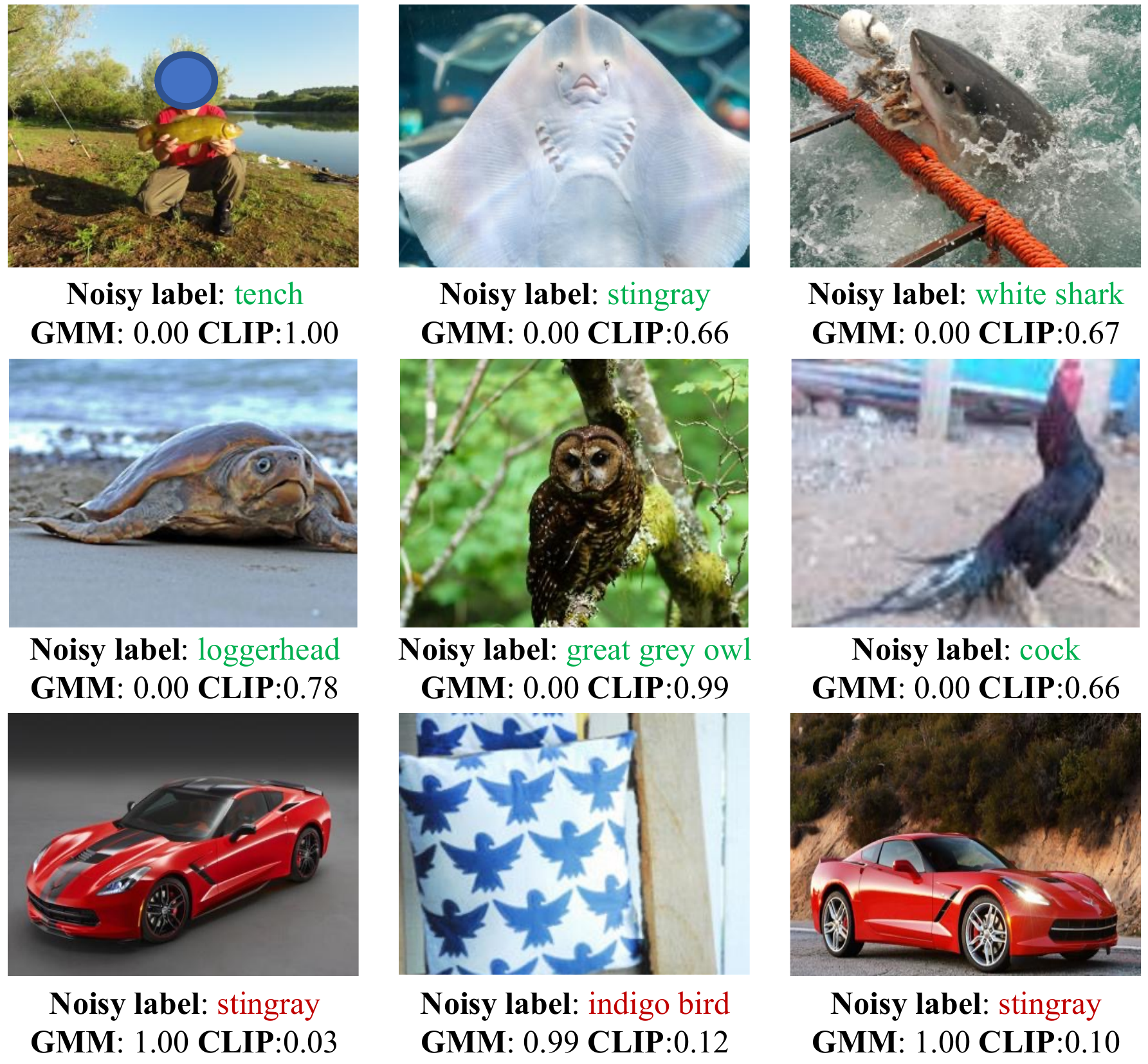}
        \caption{Examples of the selected or filtered images by CLIP on WebVision. The first two rows show the selected clean samples. The last row shows those filtered noisy samples. The confidences from GMM and CLIP are presented at the bottom of the image. Here, \textcolor[RGB]{192,0,0}{red} denotes the annotated noisy label is wrong and \textcolor[RGB]{0,176,80}{green} represents the annotated noisy label is consistent with its true label.}
	\label{fig:vis_sample}
\end{figure}

\begin{figure*}[t!]
	\centering
	\includegraphics[width=1.0\linewidth]{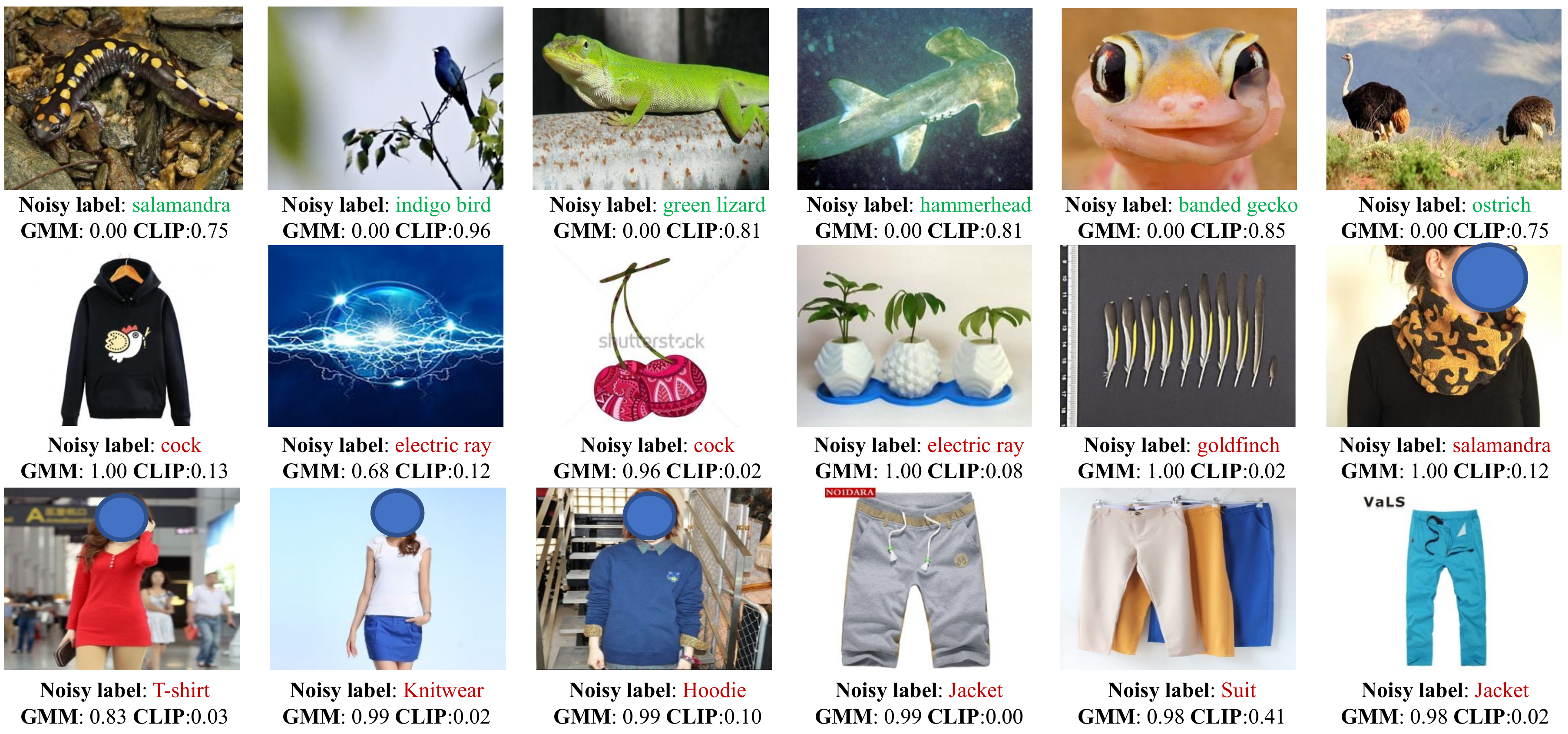}
        \caption{More examples of the selected or filtered images by CLIP. The first two rows are from WebVision and the last row is from Clothing1M. The confidence from GMM and CLIP is presented at the bottom of the image, respectively. Here, \textcolor[RGB]{192,0,0}{red} denotes the annotated noisy label is wrong and \textcolor[RGB]{0,176,80}{green} represents the annotated noisy label is consistent with its true label.}
	\label{fig:supp_webvision}
\end{figure*}

\noindent\textbf{Effect of $\delta$ in noise-aware margin regularization.} To understand the influence of the noise-aware margin, we vary $\delta$ from 0.0 to 1.0 and keep other hyperparameters fixed. As shown in Figure~\ref{fig:ab_margin}, small $\delta$~(\eg, $\delta = 0.1$) leads to higher top-1 and top-5 accuracy both on WebVision~\citep{li2017webvision} and ImageNet~\citep{deng2009imagenet}. The performance remains relatively stable when $\delta \leq 0.5$. However, when $\delta$ is pretty large~(\eg, $\delta = 1.0$), the accuracy drops considerably. The top-1 accuracy on WebVision is only 31.0\%. These results support our motivation that the noise-aware margin plays a role in preventing overconfidence effect. This regularization can provide robustness to label noise. Nevertheless, note that too large $\delta$ will hinder the optimization of the classifier, which results in poor performance. Empirically, small $\delta$ is recommended.

\noindent\textbf{Training time analysis.} We analyze the training time of our proposed method on WebVision to understand its efficiency. The experiment is conducted on a single NVIDIA 3090 GPU. For CLIP-based sample selection, it takes around 73.9 seconds with a batch size of 1000. After the pretraining stage, our model is trained for just under 40 minutes.

\subsection{Visualization}
Figure~\ref{fig:distribution} and~\ref{fig:clothing_distribution} compares the confidence distribution between GMM~\citep{Li2020DivideMix:} and CLIP~\citep{radford2021learning} on WebVision and Clothing1M. It can be seen that the confidence from GMM is near 0 or 1 after training while the distribution of CLIP is much more scattered. It indicates that some noisy samples have been memorized by DNNs and the confidence is less discriminative for distinguishing the clean and noisy samples. In Figure~\ref{fig:vis_sample}, we show several selected or filtered training images from the WebVision dataset. We compare the selection between CLIP and GMM. As we can see, CLIP can identify some clean samples which are regarded as noisy samples by GMM. 
Meanwhile, CLIP can also filter out some noisy samples that GMM fails to find, based on the predicted confidences on noisy labels. These results support our assumption that some noisy samples are memorized by DNNs and cannot be filtered based on the small loss criterion. These samples often share similar visual patterns. For instance, the pillow image~(Row 3, Column 2 in Figure~\ref{fig:vis_sample}) with a repeated bird pattern is mistakenly identified as an actual bird by the GMM. In contrast, CLIP's prior knowledge helps mitigate this issue. Benefiting from pretrained on large-scale web data, the CLIP model can help detect more noisy samples with its powerful zero-shot capability. More examples are presented in Figure~\ref{fig:supp_webvision}.

\section{Conclusion}
\label{sec:conclution}
In this paper, we propose to leverage the vision-language pretrained surrogate model CLIP to help select clean samples when dealing with noisy labels. Born with the powerful capability of zero-shot inference, CLIP can identify some noisy samples memorized by deep neural networks, based on the predicted confidences on noisy labels. Furthermore, our noise-aware balanced margin adaptive loss facilitates the learning of the classifier, which can mitigate the introduced selection bias from CLIP. The significant improvement on multiple noisy datasets verifies the effectiveness of our method without CLIP involved at the inference stage.


%
\section*{Conflict of interest}
The authors declare that they have no conflict of interest.

\section*{Data availability}
The datasets analyzed during the current study are available in \url{https://www.image-net.org/},
\url{https://www.cs.toronto.edu/~kriz/cifar.html}, \url{http://noisylabels.com/},  \url{https://google.github.io/controlled-noisy-web-labels/} and \url{https://data.vision.ee.ethz.ch/cvl/webvision/dataset2017.html}.
No new datasets were generated.

\bibliographystyle{spbasic}      
\bibliography{ref}   

%
%

\end{document}